\documentclass[letterpaper, 10 pt, conference]{ieeeconf}  

\IEEEoverridecommandlockouts                                                                                        
\usepackage{graphicx}
\usepackage{epsfig}

\usepackage{times}
\usepackage{amsmath}
\usepackage{amssymb}

\usepackage{booktabs}
\usepackage{tabularx}
\usepackage{array}
\usepackage{makecell}
\usepackage{pifont}
\usepackage{multirow}
\usepackage[table]{xcolor}
\let\labelindent\relax
\usepackage{enumitem}

\usepackage{algorithm}
\usepackage{algpseudocode}

\usepackage{cite}
\makeatletter
\def\endfigure{\end@float}
\def\endtable{\end@float}
\makeatother
\usepackage[caption=false, font=footnotesize, subrefformat=parens, listofformat=parens]{subfig}
\usepackage{placeins}
\usepackage{xcolor}
\makeatletter
\let\NAT@parse\undefined
\makeatother

\usepackage[
    colorlinks=true,
    linkcolor=blue,
    citecolor=blue,
    urlcolor=blue
]{hyperref}

\DeclareRobustCommand{\AlgCallRef}[2]{  \hyperref[#1]{\textcolor{blue}{\ensuremath{\mathrm{#2}}}}}

\newcolumntype{Y}{>{\centering\arraybackslash}X}
\newcolumntype{L}[1]{>{\raggedright\arraybackslash}p{#1}}
\newcolumntype{C}[1]{>{\centering\arraybackslash}p{#1}}

\algrenewcommand\algorithmicrequire{\textbf{Input:}}
\algrenewcommand\algorithmicensure{\textbf{Output:}}

\DeclareMathOperator*{\argminB}{argmin}   
\title{\LARGE \bf
PATACON: GPU-Parallel Tangent-Bundle RRT-Connect for Constrained Motion Planning
}

\author{
Gahyun Oh$^{1}$,
Junho An$^{1}$,
Suhyun Jeon$^{2}$,
Suhan Park$^{1,*}$\thanks{$^{1}$Department of Robotics, Kwangwoon University,
Seoul, Republic of Korea.}\thanks{$^{2}$Department of Intelligence and Information,
Graduate School of Convergence Science and Technology,
Seoul National University, Republic of Korea.}\thanks{$^{*}$Corresponding author.}
}

\begin{document}

\maketitle
\thispagestyle{empty}
\pagestyle{empty}

\begin{abstract}
Fast whole-body motion planning enables continuous replanning for humanoid robots in dynamic environments while satisfying collision avoidance and kinematic constraints. However, constrained motion planning incurs substantial computational cost due to repeated constraint projection in addition to collision checking and nearest-neighbor search. Recently, GPU-based massively parallel planners have been proposed to reduce computational cost. Nevertheless, these planners still steer in the ambient configuration space, so projecting candidates that deviate from the manifold remains a significant source of overhead, particularly in highly constrained, high-dimensional problems. In this work, we introduce PATACON, a GPU-parallel bidirectional planner that explores tangent space to exploit the local manifold geometry induced by known constraint functions. Experiments on dual-arm closed-chain and humanoid whole-body constrained planning tasks show that PATACON outperforms the prior work as configuration-space dimension and constraint codimension grow, achieving up to an 11.25 times speedup. These results suggest that GPU-parallel tangent-space exploration brings constrained whole-body motion planning closer to real-time performance.

\end{abstract}

\section{Introduction}
\label{sec:introduction}

Dual-arm closed-chain and humanoid whole-body motion require maintaining task-dependent kinematic relationships while satisfying joint limits and avoiding collisions. These kinematic requirements can be formulated as nonlinear equality constraints, which define a lower-dimensional constraint manifold embedded in a high-dimensional configuration space~\cite{KingstonConstraintSurvey}. In sampling-based motion planning, collision checking and nearest-neighbor queries are generally major sources of computational cost\cite{YershovaNN,KleinbortBottleneck}. Planning on a constraint manifold introduces additional overhead because constraint projection is performed repeatedly throughout tree expansion. Reducing this computational burden is particularly important in scenarios that require repeated replanning as the target, obstacles, or task conditions change, where low planning latency is essential for responsive execution.

\begin{figure}[t]
    \centering
    \includegraphics[
        width=\columnwidth,
        trim=0 0 0 0,
        clip
    ]{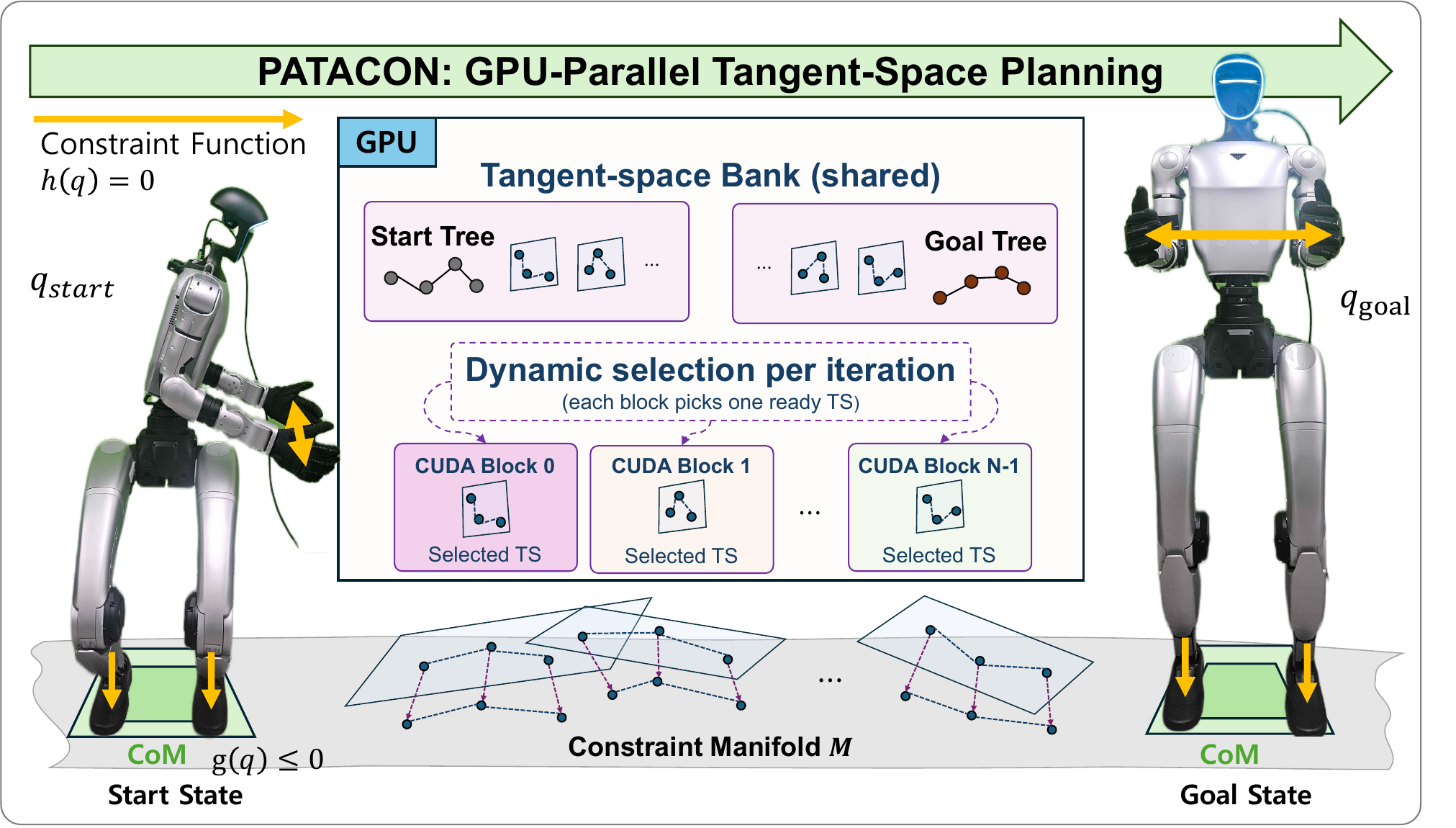}
    \vspace{-7mm}
    \caption{
    Overview of PATACON, a GPU-parallel bidirectional constrained planner. Task-specific constraints define a constraint manifold, and CUDA blocks select local tangent spaces from shared tree-specific banks to explore different regions of the manifold concurrently. Candidate motion segments generated in each tangent space are projected and validated on the manifold before insertion into the trees.
        }
    \label{fig:overview}
    \vspace{-4mm}
\end{figure}

A widely used approach generates candidate motions in the ambient configuration space and projects them onto the manifold as the tree grows~\cite{KingstonConstraintSurvey}. cpRRTC~\cite{cpRRTC} brought this approach to the GPU by assigning each thread block to steer from the nearest node toward a random sample, project the waypoints of the resulting straight-line segment in parallel, and perform collision checks with early termination. This reduces constrained planning to milliseconds, but the exploration scheme is unchanged; each extension still steers in the ambient space and relies on projection to satisfy the constraints.

Ambient-space steering, however, ignores the shape of the manifold. Kim et al.~\cite{TBRRT} showed that a target sampled in the ambient space can contribute little progress along the manifold once projected, and that the resulting large number of projections degrades both computational efficiency and the Voronoi bias of RRT. Because the constraint Jacobian provides the tangent space at any configuration, displacements can instead be restricted to it: Stilman~\cite{StilmanTaskConstraint} observed that tangent displacements are more likely to remain within the constraint tolerance, and TB-RRT~\cite{TBRRT} grows trees on tangent spaces bounded by the local curvature, outperforming projection-based planners with larger gains on more complex problems. However, TB-RRT expands one node at a time, incurs additional preprocessing and bookkeeping for its tangent spaces, and relies on lazy projection. Therefore, the feasibility checked on tangent spaces must be re-evaluated after projection.

We propose PATACON (\textbf{PA}rallel \textbf{TA}ngent-bundle RRT-\textbf{CON}nect), a GPU-parallel bi-directional planner that brings tangent-space-guided exploration to massively parallel tree expansion, as illustrated in Fig.~\ref{fig:overview}. 
Unlike TB-RRT, PATACON projects and validates every candidate before tree insertion, so every node is feasible. Its efficiency has two sources: tangent-space candidates are more likely to remain within the constraint tolerance~\cite{StilmanTaskConstraint} and avoid ambient targets whose projections make little progress along the manifold~\cite{TBRRT}, and a single sample and nearest-neighbor query yield multiple candidate edges that are validated concurrently on the GPU, which amortizes the preprocessing and bookkeeping of tangent spaces over multiple edges. 

Our main contributions are as follows:

\begin{enumerate}[leftmargin=*]
    \item \textbf{Tangent-space-guided GPU-parallel extension.} PATACON replaces the ambient-space steering of existing GPU-parallel planners with extension along local tangent spaces, so that candidates follow the manifold geometry before projection.

    \item \textbf{Shared tangent-space banks.} Tangent spaces are maintained
    in tree-specific banks shared across thread blocks, together with the
    correspondence between projected manifold states and their tangent-space
    coordinates, so that multiple blocks can explore different regions of the
    manifold concurrently.

    \item \textbf{Speculative multi-edge expansion.} From one sample and one nearest-neighbor query, PATACON generates consecutive candidate edges along a tangent direction, validates them concurrently, and inserts the longest valid prefix, unlike the step-by-step CONNECT heuristic of RRT-Connect~\cite{RRTConnect}.

\end{enumerate}

We evaluate PATACON on eight benchmarks ranging from 7-DoF manipulation to 35-DoF humanoid whole-body planning. Although cpRRTC is faster on low-dimensional problems, PATACON shows increasing advantages with configuration-space dimension and constraint restrictiveness, reaching a $11.25\times$ speedup with a 100\% success rate.

\section{Related Work}
\label{sec:related_work}

Sampling-based constrained motion planners enforce nonlinear equality constraints through numerical projection or local approximations of the constraint manifold~\cite{KingstonConstraintSurvey}. 
Projection-based methods range
from randomized gradient descent for closed chains~\cite{YakeyClosedChain}
to Jacobian-based tangent-space sampling and first-order
retraction~\cite{StilmanTaskConstraint} and CBiRRT~\cite{CBiRRT} with Task
Space Regions~\cite{TaskSpaceRegions}, whereas AtlasRRT~\cite{AtlasRRT}
and TB-RRT~\cite{TBRRT} grow trees on local charts or tangent spaces, and
Kingston et al.~\cite{KingstonIMACS} unify these mechanisms. These methods,
however, are primarily sequential.

Learning-based planners reduce online search through learned latent
spaces~\cite{Ichter2019Latent,Eqnet2021,ParkLatentMotion} or constraint-aware
samplers~\cite{CoMPNetX}, but require offline data generation and
training; we instead accelerate planning directly from the given
constraint functions.

Parallelization reduces the wall-clock cost of planning through GPU
collision checking~\cite{BialkowskiParallelRRT,PanGPUCollision}, multiple
or shared trees on CPUs~\cite{CForest,IchnowskiParallel,JacobsParallelSBMP},
SIMD vectorization~\cite{VAMP,McVAMP}, and GPU trajectory
optimization~\cite{cuRobo}. pRRTC~\cite{pRRTC} and BatchRRT~\cite{BatchRRT}
run RRT-style tree growth on the GPU, and cpRRTC~\cite{cpRRTC} extends
pRRTC to constrained planning by projecting motion-segment waypoints in
parallel. However, the steering methods remain in the ambient space, so
parallelism lowers the cost of each projection but not the off-manifold
motion to be projected.

Simply adding tangent-space steering to such planners is insufficient,
because tangent-space management and local nearest-neighbor search add
per-extension overhead. PATACON instead exploits the fact that
tangent-space candidates stay close to the manifold: one sample and
nearest-neighbor query yield several consecutive candidate edges, which
are validated in parallel on the GPU to amortize this overhead.

\section{Problem Definition}
\label{sec:problem_definition}

Let $\mathcal{C}\subseteq\mathbb{R}^{n}$ denote the joint-feasible robot configuration space, where $q\in\mathcal{C}$ represents a robot configuration. Task-dependent kinematic constraints are represented by a differentiable nonlinear equality-constraint function $h:\mathcal{C}\rightarrow\mathbb{R}^{m}$, which defines the constraint manifold as
\vspace{0mm}
\begin{equation}
    \mathcal{M}
    =
    \{q\in\mathcal{C}\mid h(q)=0\}.
    \label{eq:constraint_manifold}
\end{equation}

At a regular configuration $q$ in the interior of $\mathcal{C}$, we assume that the constraint Jacobian
$J_h(q)=\partial h(q)/\partial q$ has full row rank $m$. The constraint manifold is then locally of dimension $d=n-m$, and its tangent space is given by
\vspace{0mm}
\begin{equation}
    T_q\mathcal{M}
    =
    \{v\in\mathbb{R}^{n}\mid J_h(q)v=0\}.
    \label{eq:tangent_space}
\end{equation}

Equivalently, let $N(q)\in\mathbb{R}^{n\times d}$ denote an orthonormal basis of the null space of $J_h(q)$, such that $J_h(q)N(q)=0$ and $N(q)^{\top}N(q)=I_d$. This basis provides the local tangent-space representation used by the planner.

Let $\mathcal{C}_{\mathrm{free}}\subseteq\mathcal{C}$ denote the set of configurations that are free of both self-collision and environmental collision. The feasible portion of the constraint manifold is then
\vspace{0mm}
\begin{equation}
    \mathcal{M}_{\mathrm{free}}
    =
    \mathcal{M}\cap\mathcal{C}_{\mathrm{free}}.
    \label{eq:feasible_manifold}
\end{equation}

Given feasible start and goal configurations
$q_{\mathrm{start}},q_{\mathrm{goal}}\in\mathcal{M}_{\mathrm{free}}$,
the constrained motion-planning problem considered in this work is to find a continuous path
\vspace{0mm}
\begin{equation}
    \sigma:[0,1]\rightarrow\mathcal{M}_{\mathrm{free}},
    \label{eq:constrained_path}
\end{equation}

such that
\vspace{0mm}
\begin{equation}
    \sigma(0)=q_{\mathrm{start}},
    \qquad
    \sigma(1)=q_{\mathrm{goal}}.
    \label{eq:path_boundary}
\end{equation}

Thus, every configuration along the resulting path must satisfy the nonlinear kinematic constraints, joint limits, and collision-avoidance requirements.

\section{GPU-Parallel Tangent-Bundle RRT-Connect}
\label{sec:method}

PATACON restructures tangent-space-guided RRT-Connect for GPU-parallel execution through two algorithmic components: \textbf{Tangent-Guided Sampling (TGS)} and \textbf{Tangent-Guided Multi-Edge Projection (TGMP)}. TGS guides exploration using local tangent-space geometry, while TGMP converts each TGS exploration decision into a multi-step expansion that generates multiple consecutive candidate edges. PATACON combines concurrent exploration across multiple tangent-space regions with GPU-parallel processing of the candidate edges generated by TGMP.

\begin{figure}[t]
    \centering
    \includegraphics[
    width=\columnwidth,
    trim=175 150 150 150,
    clip
    ]{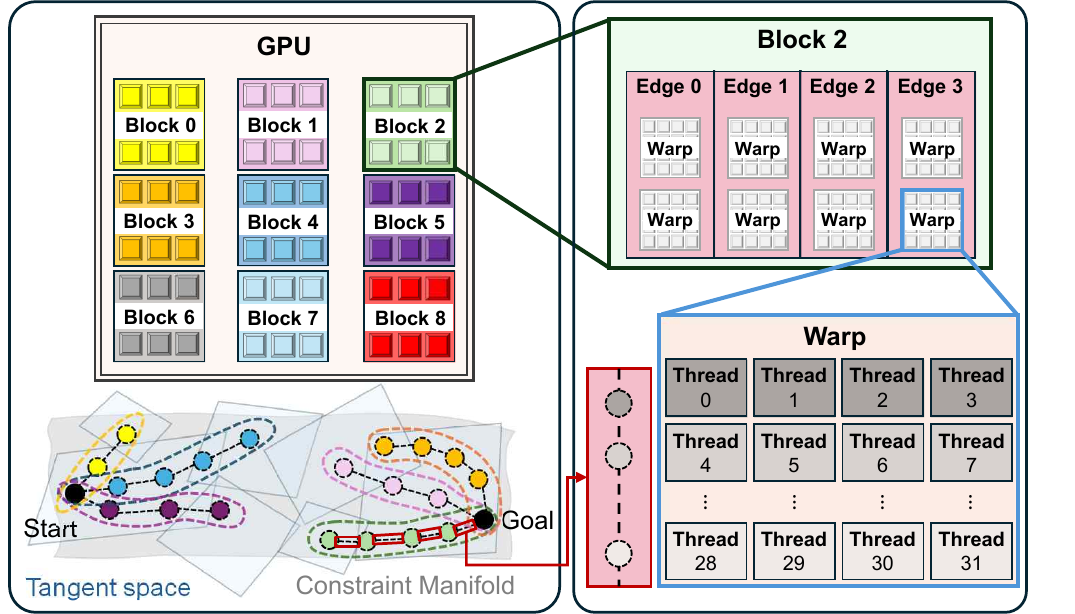}
    \caption{GPU mapping of PATACON. CUDA blocks concurrently explore local tangent-space regions, while candidate edges generated by each EXTEND are distributed across cooperative thread groups for parallel validation.}
    \label{fig:gpu_structure}
    \vspace{-4mm}
\end{figure}

In each exploration iteration, PATACON expands the selected tree through the TGS--TGMP pipeline, projects and validates the resulting candidate edges in parallel, and inserts only the longest consecutively valid prefix. A successful EXTEND is followed by a target-directed CONNECT toward the opposite tree. The following subsections describe these stages in order. Fig.~\ref{fig:gpu_structure} illustrates the GPU execution structure, while Algorithm~\ref{alg:patacon} summarizes the overall planning procedure.

\subsection{Tangent-Guided Sampling for Concurrent Exploration}
\label{sec:concurrent_exploration}

TGS determines the local region and expansion state used for each tree-expansion attempt. For a tangent-space root $q_r\in\mathcal{M}$, PATACON uses the orthonormal null-space basis $N(q_r)$ defined in Sec.~\ref{sec:problem_definition}. The corresponding tangent-space projector is

\begin{equation}
P_{\mathcal{T}_r}
=
N(q_r)N(q_r)^{\top}.
\label{eq:tangent_projector}
\end{equation}

\begin{equation}
\mathcal{T}_r
=
\left\{
q_r + N(q_r)c
\mid
c \in \mathbb{R}^{d}
\right\}.
\label{eq:affine_tangent_space}
\end{equation}

Sampling within $\mathcal{T}_r$ guides exploration toward directions consistent with the local constraint geometry, reducing unnecessary exploration away from the manifold before projection. This geometric guidance, however, introduces additional computation for tangent-space management, selection, and local nearest-neighbor search, which motivates the multi-step expansion of TGMP to amortize this overhead.
Each tree node stores its projected manifold configuration $q_i \in \mathcal{M}_{\mathrm{free}}$, parent index, associated tangent-space index $\tau(q_i)$, and local tangent-space representation $\bar q_i \in \mathcal{T}_{\tau(q_i)}$. The projected configuration $q_i$ is used for feasibility and path validation, whereas $\bar q_i$ is used for tangent-space exploration and tangent-space-local nearest-neighbor search.

PATACON maintains separate tangent-space banks in GPU global memory for the start and goal trees, with each bank shared across the CUDA blocks exploring the corresponding tree. At initialization, tangent spaces rooted at $q_{\mathrm{start}}$ and $q_{\mathrm{goal}}$ are inserted into their respective banks. A newly generated tangent space becomes selectable only after its root node and associated tangent-space data have been fully initialized.

At each exploration iteration, a CUDA block selects the grow tree $T_{\mathrm{grow}}$ and opposite tree $T_{\mathrm{opp}}$ according to their current sizes. When the two trees have comparable sizes, the grow tree is alternated; when their size difference exceeds a prescribed threshold, the smaller tree is preferentially expanded until the imbalance is reduced, following the balanced bidirectional search principle~\cite{BalancedRRT}.

After selecting the grow tree, each CUDA block dynamically selects a ready tangent space from the corresponding bank. The initial bank index varies with the block and planning iteration, after which the bank is scanned cyclically until a ready tangent space is found. Therefore, tangent spaces are dynamically selected rather than statically assigned to individual CUDA blocks.

Following TB-RRT~\cite{TBRRT}, PATACON applies backtracking prevention when sampling from a newly created tangent space. Let \(q_p\) denote the root of its parent tangent space and \(q_r\) the root of the new tangent space. The displacement \(q_r-q_p\) is projected onto the new tangent space to obtain a forward direction \(d_r\). If the sampled tangent-space displacement has a negative inner product with \(d_r\), its direction is reversed. This restricts sampling to the forward half-space of the tangent space and prevents the tree from backtracking toward previously explored regions.

Within the selected tangent space $\mathcal{T}_r$, a block-local Halton sample~\cite{Halton1960,BranickyQuasiRandom} determines a normalized combination of tangent-basis directions and a displacement that remains within the joint-space sampling bounds, producing a tangent-space sample $\bar q_{\mathrm{rand}}$. Let $V(T,r)$ denote the set of configurations in tree $T$ associated with tangent space $\mathcal{T}_r$. The expansion node is then selected by a nearest-neighbor query over $V(T_{\mathrm{grow}},r)$,

\vspace{-4mm}

\begin{equation}
q_{\mathrm{near}}
=
\argminB_{q_i \in V(T_{\mathrm{grow}},r)}
\operatorname{dist}
\left(
\bar q_i,
\bar q_{\mathrm{rand}}
\right),
\label{eq:tangent_nn}
\end{equation}

where $\bar q_i$ denotes the tangent-space representation of $q_i$ in $\mathcal{T}_r$, and $\operatorname{dist}(q_a,q_b)=\|q_a-q_b\|_2$. This nearest-neighbor search is parallelized within each CUDA block using thread-level distance evaluation followed by block-level reduction. The sample $\bar q_{\mathrm{rand}}$ and the tangent-space representation $\bar q_{\mathrm{near}}$ of the selected node $q_{\mathrm{near}}$ are then passed to TGMP for subsequent expansion.

\begin{algorithm}[t]
\caption{PATACON
$(q_{\mathrm{start}},q_{\mathrm{goal}},N_{\mathrm{blk}},K_{\max})$}
\label{alg:patacon}
\footnotesize
\begin{algorithmic}[1]

\State $T_s.\mathrm{AddVertex}(q_{\mathrm{start}})$,
       $T_g.\mathrm{AddVertex}(q_{\mathrm{goal}})$

\State $\mathrm{TSBank}_s.\mathrm{Add}
       (\mathrm{CreateTangentSpace}(q_{\mathrm{start}}))$

\State $\mathrm{TSBank}_g.\mathrm{Add}
       (\mathrm{CreateTangentSpace}(q_{\mathrm{goal}}))$

\State \textbf{for each CUDA block}
$b=1,\ldots,N_{\mathrm{blk}}$
\textbf{ in parallel}

\For{$i=1,\ldots,I_{\max}$}

    \State $(T_{\mathrm{grow}},T_{\mathrm{opp}})
    \gets
    \Call{BalanceTrees}{T_s,T_g,b,i}$

    \State $\mathcal{T}_r
    \gets
    \Call{SelectTangentSpace}
    {\mathrm{TSBank}(T_{\mathrm{grow}}),b,i}$

    \State $\bar{q}_{\mathrm{rand}}
    \gets
    \Call{Sample}{\mathcal{T}_r}$

    \State $(q_{\mathrm{near}},\bar{q}_{\mathrm{near}})
    \gets
    \Call{NearestInTangentSpace}
    {T_{\mathrm{grow}},
     \mathcal{T}_r,
     \bar{q}_{\mathrm{rand}}}$

    \State $v_{\mathrm{ext}}
    \gets
    \bar{q}_{\mathrm{rand}}
    -
    \bar{q}_{\mathrm{near}}$

    \State $(s_{\mathrm{ext}},q_{\mathrm{end}})
    \gets
    \hyperref[alg:ts_extend]
    {\textcolor{blue}{\textsc{Extend}}}
    (T_{\mathrm{grow}},
      q_{\mathrm{near}},
      \bar{q}_{\mathrm{near}},
      v_{\mathrm{ext}},
      K_{\max})$

    \If{$s_{\mathrm{ext}}=\mathrm{Trapped}$}
        \Comment{\textcolor{gray}{No valid extension}}
        \State \textbf{continue}
    \EndIf

    \State $q_{\mathrm{tar}}
    \gets
    \displaystyle
    \arg\min_{q_j\in V(T_{\mathrm{opp}})}
    \operatorname{dist}
    (q_j,q_{\mathrm{end}})$

    \State $(s_{\mathrm{conn}},\sigma)
    \gets
    \hyperref[alg:check_connection]
    {\textcolor{blue}{\textsc{CheckConnection}}}
    (T_{\mathrm{grow}},
      T_{\mathrm{opp}},
      q_{\mathrm{end}},
      q_{\mathrm{tar}})$

    \If{$s_{\mathrm{conn}}=\mathrm{Connected}$}
        \State \Return $\sigma$
    \EndIf

    \State $q_{\mathrm{cur}}
    \gets q_{\mathrm{end}}$

    \For{$j=1,\ldots,J_{\max}$}
        \Comment{\textcolor{gray}{CONNECT}}

        \State $\mathcal{T}_r
        \gets
        \Call{TangentSpace}{q_{\mathrm{cur}}}$

        \State $\bar{q}_{\mathrm{cur}}
        \gets
        \Call{TangentConfig}{q_{\mathrm{cur}}}$

        \State $v_{\mathrm{con}}
        \gets
        P_{\mathcal{T}_r}
        (q_{\mathrm{tar}}-q_{\mathrm{cur}})$

        \State $(s_{\mathrm{con}},q_{\mathrm{end}})
        \gets
        \hyperref[alg:ts_extend]
        {\textcolor{blue}{\textsc{Extend}}}
        \bigl(
        T_{\mathrm{grow}},
        q_{\mathrm{cur}},
        \bar{q}_{\mathrm{cur}},
        v_{\mathrm{con}},
        K_{\max}
        \bigr)$

        \If{$s_{\mathrm{con}}=\mathrm{Trapped}$}
            \Comment{\textcolor{gray}{No valid extension}}
            \State \textbf{break}
        \EndIf

        \State $q_{\mathrm{cur}}
        \gets q_{\mathrm{end}}$

        \State $(s_{\mathrm{conn}},\sigma)
        \gets
        \hyperref[alg:check_connection]
        {\textcolor{blue}{\textsc{CheckConnection}}}
        (T_{\mathrm{grow}},
          T_{\mathrm{opp}},
          q_{\mathrm{cur}},
          q_{\mathrm{tar}})$

        \If{$s_{\mathrm{conn}}=\mathrm{Connected}$}
            \State \Return $\sigma$
        \EndIf

    \EndFor

\EndFor

\State \Return $\mathrm{Invalid}$

\end{algorithmic}
\end{algorithm}

\subsection{Tangent-Guided Multi-Edge Projection}
\label{sec:tgmp}

TGMP converts the expansion generated by TGS into a multi-step tree expansion. Instead of generating only one candidate edge per sampling and nearest-neighbor operation, TGMP reuses the selected tangent-space context and expansion direction to generate multiple consecutive candidates within a single EXTEND call.

For exploratory expansion, the tangent-space sample $\bar q_{\mathrm{rand}}$ and the local representation $\bar q_{\mathrm{near}}$ of the selected expansion node define the displacement

\begin{equation}
v_{\mathrm{ext}}
=
\bar q_{\mathrm{rand}}
-
\bar q_{\mathrm{near}}.
\label{eq:extend_direction}
\end{equation}

If $\|v_{\mathrm{ext}}\|_2 \leq \epsilon_{\mathrm{dir}}$, EXTEND returns \textsc{Trapped}. Otherwise, the displacement is normalized as

\begin{equation}
u
=
\frac{v_{\mathrm{ext}}}
{\|v_{\mathrm{ext}}\|_2},
\label{eq:normalized_direction}
\end{equation}

and generates the $k$-th tangent-space candidate as

\begin{equation}
\bar q_k
=
\bar q_{\mathrm{near}}
+
k\xi u,
\qquad
k=1,\ldots,K,
\label{eq:tangent_candidate}
\end{equation}

where $\xi$ denotes the extension step size and $K\leq K_{\max}$. The ordered sequence $Q_{\mathrm{cand}}=(\bar q_1,\ldots,\bar q_K)$ represents a multi-step expansion along the common tangent-space direction and defines the candidate edges processed in the subsequent projection and validation stage.

\begin{equation}
E_M(\bar q_k)
=
\|h(\bar q_k)\|_2 .
\label{eq:tangent_error}
\end{equation}

\begin{figure*}[t]
\centering
\includegraphics[
width=\textwidth,
trim=0 380 0 70,
clip
]{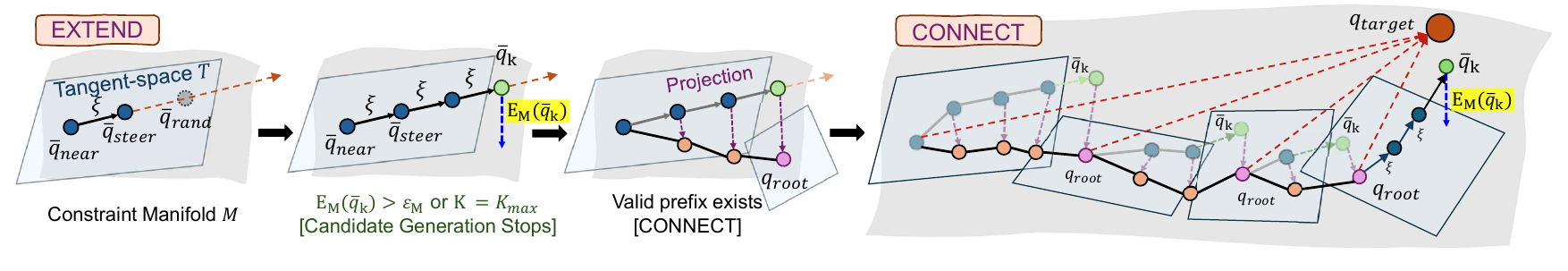}
\caption{Tangent-Space EXTEND and CONNECT. EXTEND generates multiple candidates along a common tangent direction, while CONNECT repeatedly applies EXTEND toward a fixed target in the opposite tree.}
\label{fig:extend_connect}
\end{figure*}

Because the affine tangent space provides only a local linear approximation of the nonlinear constraint manifold, a generated candidate $\bar{q}_k$ does not necessarily satisfy the nonlinear constraints. PATACON evaluates its constraint residual as

Candidate generation terminates when $K_{\max}$ candidates have been generated or immediately after appending the first candidate satisfying

\begin{equation}
E_M(\bar q_k)
>
\epsilon_M.
\label{eq:tangent_error_threshold}
\end{equation}

The threshold-crossing candidate remains in $Q_{\mathrm{cand}}$ and is passed to the projection and validation stage. The residual $E_M$ is used only as a practical criterion for determining when the current tangent-space approximation should be renewed and is not interpreted as the geometric distance from $\bar q_k$ to the constraint manifold. Moreover, generating a threshold-crossing candidate alone does not establish its feasibility or immediately create a new tangent space.

TGMP itself defines the multi-step expansion independently of the execution schedule used for candidate-edge validation. Therefore, the candidate edges generated by TGMP can be processed sequentially, as in the $+\mathrm{TGMP}$ ablation variant, or concurrently on the GPU, as in the full PATACON planner.

\begin{algorithm}[t]
\caption{EXTEND
$(T,q_{\mathrm{cur}},\bar{q}_{\mathrm{cur}},v,K_{\max})$}
\label{alg:ts_extend}
\footnotesize
\begin{algorithmic}[1]

\If{$\|v\|_2\leq\epsilon_{\mathrm{dir}}$}
    \Comment{\textcolor{gray}{No extension direction}}
    \State \Return $(\mathrm{Trapped},\varnothing)$
\EndIf

\State $u
\gets
v/\|v\|_2$

\State $\mathcal{Q}_{\mathrm{cand}}
\gets
\emptyset$
\Comment{\textcolor{gray}{Tangent-space candidate batch}}

\For{$k=1,\ldots,K_{\max}$}
    \Comment{\textcolor{gray}{Generate tangent candidates}}

    \State $\bar{q}_k
    \gets
    \bar{q}_{\mathrm{cur}}
    +
    k\xi u$

    \State Append $\bar{q}_k$
    to $\mathcal{Q}_{\mathrm{cand}}$

    \If{$E_M(\bar{q}_k)>\epsilon_M$}
        \Comment{\textcolor{gray}{Check tangent error}}
        \State \textbf{break}
    \EndIf

\EndFor

\State $\mathcal{Q}_{\mathrm{valid}}
\gets
\Call{ProjectAndCollisionCheck}
{q_{\mathrm{cur}},
 \mathcal{Q}_{\mathrm{cand}}}$

\If{$|\mathcal{Q}_{\mathrm{valid}}|=0$}
    \Comment{\textcolor{gray}{No valid prefix}}
    \State \Return $(\mathrm{Trapped},\varnothing)$
\EndIf

\State $r
\gets
\Call{TangentSpaceIndex}
{q_{\mathrm{cur}}}$

\State $q_{\mathrm{end}}
\gets
\Call{InsertBatch}
{T,
 q_{\mathrm{cur}},
 \mathcal{Q}_{\mathrm{valid}},
 r}$

\If{$|\mathcal{Q}_{\mathrm{valid}}|
=
|\mathcal{Q}_{\mathrm{cand}}|$}
    \Comment{\textcolor{gray}{Full batch for TS renewal}}

    \State $\bar{q}_{\mathrm{end}}
    \gets
    \Call{Last}{\mathcal{Q}_{\mathrm{cand}}}$

    \If{$E_M(\bar{q}_{\mathrm{end}})>\epsilon_M$}
        \Comment{\textcolor{gray}{Check TS renewal}}
        \State $\mathrm{CreateTangentSpace}(T,q_{\mathrm{end}})$
    \EndIf

\EndIf

\State \Return
$(\mathrm{Advanced},q_{\mathrm{end}})$

\end{algorithmic}
\end{algorithm}

\subsection{GPU-Parallel Edge Projection and Validation}
\label{sec:parallel_validation}

PATACON parallelizes the projection and validation of the multiple candidate edges generated by TGMP. Different candidate edges are assigned to separate cooperative thread groups and processed concurrently, whereas the waypoint projection within each individual edge follows the original sequential projection order.

For each candidate edge, intermediate waypoints are generated according to the edge-discretization resolution $G$. Starting from the edge source, the waypoints are projected sequentially toward the constraint manifold so that each projection step follows the preceding projected waypoint. Collision checking and feasibility validation are performed for the projected edge, while multiple candidate edges undergo this procedure concurrently on the GPU.

An edge is considered valid only when its sequential waypoint projection converges within the prescribed projection tolerances and all projected configurations satisfy the joint limits and collision constraints. During target-directed CONNECT, the target-progress condition described in Sec.~\ref{sec:connect} is additionally applied.

Although candidate edges are processed concurrently, their validation results are consumed in the original TGMP expansion order. Only the longest consecutively valid prefix beginning with the first candidate edge is inserted into the tree. If the first candidate edge is invalid, EXTEND returns \textsc{Trapped}; otherwise, the valid prefix is inserted and EXTEND returns \textsc{Advanced} with the final inserted configuration $q_{\mathrm{end}}$.

Tangent-space renewal is determined separately from valid-prefix insertion. A new tangent space is created at $q_{\mathrm{end}}$ only when the entire TGMP candidate batch is accepted and the final tangent-space candidate satisfies

\begin{equation}
E_M(\bar q_{\mathrm{end}})
>
\epsilon_M.
\label{eq:tangent_renewal}
\end{equation}

The boundary node then becomes the root of the new tangent space with $\bar q_i=q_i$; otherwise, the inserted nodes remain associated with their original tangent space. Consequently, full-batch acceptance is required only for tangent-space renewal, whereas any nonempty valid prefix is sufficient to trigger the subsequent CONNECT attempt.

\subsection{Target-Directed Connection and Bridge Validation}
\label{sec:connect}

After an exploratory EXTEND returns \textsc{Advanced}, PATACON selects the closest configuration in the opposite tree as a fixed target for the current CONNECT attempt,

\begin{equation}
q_{\mathrm{tar}}
=
\argminB_{
q_j \in V(T_{\mathrm{opp}})
}
\operatorname{dist}
(q_j,q_{\mathrm{end}}).
\label{eq:connect_target}
\end{equation}

Unlike the tangent-space-local nearest-neighbor search used by TGS, this target search considers all configurations in the opposite tree. Starting from $q_{\mathrm{cur}}=q_{\mathrm{end}}$, PATACON repeatedly applies EXTEND to $T_{\mathrm{grow}}$ toward the same fixed target $q_{\mathrm{tar}}$. During a CONNECT attempt, the opposite tree provides the target configuration but is not itself expanded by that CONNECT operation.

At each CONNECT step, let $\mathcal{T}_r$ denote the tangent space associated with $q_{\mathrm{cur}}$. The ambient displacement toward the fixed target is projected onto the tangent vector space as

\begin{equation}
v_{\mathrm{con}}
=
P_{\mathcal{T}_r}
(q_{\mathrm{tar}}-q_{\mathrm{cur}}).
\label{eq:connect_direction}
\end{equation}

The resulting $v_{\mathrm{con}}$ is passed to the same TGMP-based EXTEND procedure used during exploratory expansion. Thus, CONNECT also generates a multi-step candidate sequence and processes the resulting candidate edges using the GPU-parallel edge projection and validation procedure described in Sec.~\ref{sec:parallel_validation}.

Because numerical projection can modify the motion implied by the tangent-space direction, satisfying the projection and collision constraints alone does not guarantee progress toward the fixed target. Therefore, each candidate edge considered during CONNECT must additionally satisfy

\begin{equation}
\operatorname{dist}
(q_k,q_{\mathrm{tar}})
<
\operatorname{dist}
(q_{k-1},q_{\mathrm{tar}})
-
\epsilon_{\mathrm{prog}},
\label{eq:connect_progress}
\end{equation}

where $\epsilon_{\mathrm{prog}}>0$ denotes the minimum required reduction in distance to the target. If an edge fails this condition, that edge is rejected and valid-prefix insertion terminates before it, even if its projection and collision checks succeed.

After each successful EXTEND, $q_{\mathrm{cur}}$ is updated to the returned $q_{\mathrm{end}}$, and PATACON evaluates

\begin{equation}
\operatorname{dist}
(q_{\mathrm{cur}},q_{\mathrm{tar}})
\leq
\epsilon_{\mathrm{con}}.
\label{eq:connect_proximity}
\end{equation}

When this condition is satisfied, the remaining bridge between $q_{\mathrm{cur}}$ and $q_{\mathrm{tar}}$ is projected and validated using the same sequential waypoint projection and collision-validation procedure. The two trees are considered connected only if this final bridge validation succeeds; otherwise, the CONNECT attempt continues until EXTEND returns \textsc{Trapped} or the maximum number of EXTEND calls $J_{\max}$ is reached.

Algorithm~\ref{alg:check_connection} summarizes the target-proximity test, final bridge validation, and path construction.

\begin{algorithm}[t]
\caption{CHECK CONNECTION
$(T_{\mathrm{grow}},T_{\mathrm{opp}},q,q_{\mathrm{tar}})$}
\label{alg:check_connection}
\footnotesize
\begin{algorithmic}[1]

\If{$
\operatorname{dist}(q,q_{\mathrm{tar}})
>
\epsilon_{\mathrm{con}}
$}
    \State \Return
    $(\mathrm{NotConnected},\varnothing)$
\EndIf

\State
$(s_{\mathrm{bridge}},
  \mathcal{Q}_{\mathrm{bridge}})
\gets
\Call{ValidateBridge}
{q,q_{\mathrm{tar}}}$

\If{$s_{\mathrm{bridge}}
\neq
\mathrm{Valid}$}
    \State \Return
    $(\mathrm{NotConnected},\varnothing)$
\EndIf

\State $\sigma
\gets
\Call{TracePath}
{T_{\mathrm{grow}},
 T_{\mathrm{opp}},
 \mathcal{Q}_{\mathrm{bridge}}}$

\State \Return
$(\mathrm{Connected},\sigma)$

\end{algorithmic}
\end{algorithm}

\section{Experiments}
\label{sec:experiments}

We conducted three sets of experiments to evaluate PATACON. \textbf{First}, we compare planner performance across regimes spanning increasingly restrictive constraints and higher-dimensional configuration spaces. \textbf{Second}, we perform a cumulative ablation study by progressively introducing Tangent-Guided Sampling (TGS) and Tangent-Guided Multi-Edge Projection (TGMP) to cpRRTC. \textbf{Finally}, we evaluate solution-quality improvement under a common anytime optimization framework.

\subsection{Experimental Environment}
\label{sec:methodology}

We evaluate PATACON primarily against two representative constrained motion planners: McVAMP~\cite{McVAMP} and cpRRTC~\cite{cpRRTC}, representing CPU SIMD-accelerated manifold-constrained planning and GPU-parallel projection-based RRT-Connect, respectively. We additionally include TB-RRT~\cite{TBRRT} as a tangent-space-based constrained planner on the single-arm Franka benchmark.

For McVAMP, we use its publicly available implementation. For cpRRTC, we implement the planner by extending the publicly available pRRTC codebase according to the algorithm described in the cpRRTC paper. For TB-RRT, we use the implementation provided by OMPL.

Within each benchmark regime, all planners use the same robot model, environment, start--goal pairs, joint limits, constraint definitions, feasibility tolerances, and planning timeout. For each regime, we randomly sample 100 feasible start--goal joint-state pairs, which are shared across all evaluated planners.

All experiments were conducted on a workstation equipped with an AMD Ryzen 7 9800X3D CPU, an NVIDIA GeForce RTX 5090 GPU, and 64 GB of system memory. Unless otherwise specified, PATACON used $N_{\mathrm{blk}}=512$, $K_{\max}=4$, $\epsilon_M=0.1$, and an edge-discretization resolution of $G=16$. One-time robot, environment, and device initialization was excluded from the timing measurements, while all computations required for each planning query were included.

The benchmark robots span configuration spaces of different dimensions: (A) 7-DoF for the Franka Panda, (B) 14-DoF for the dual Franka, (C) 15-DoF for the fixed-base FFW-SG2, (D) 18-DoF for the mobile-base FFW-SG2, and (E) 35-DoF for the Unitree G1. We denote the constraints as follows: BK is a 6D bimanual kinematic-chain equality constraint on the relative end-effector pose; OR is a 2D orientation equality constraint that aligns the $z$-axes while leaving rotation about the aligned axis unconstrained; and FT is a 12D equality constraint that fixes the poses of both feet (6D per foot). CoM constrains the planar $(x,y)$ center of mass to lie within the support region using inequality constraints.

\subsection{Planning Performance across Constraint Regimes} \label{sec:constraint_regime_performance}

Fig.~\ref{fig:compare algorithm} compares the planning time and success rate across the benchmark regimes. Each trial is limited to 500~s, and the RRT extension range is set to $\xi=0.4$ where applicable. TB-RRT is evaluated only on the single-arm Franka benchmark because preliminary evaluations showed substantially larger planning times on higher-dimensional problems. McVAMP is evaluated on the Franka and dual-Franka benchmarks, but is omitted from the higher-dimensional FFW-SG2 and Unitree G1 benchmarks due to similarly high computational cost. We therefore focus on cpRRTC and PATACON in these higher-dimensional regimes.

\begin{figure}[t]
    \centering
    \includegraphics[
        width=\columnwidth
                    ]{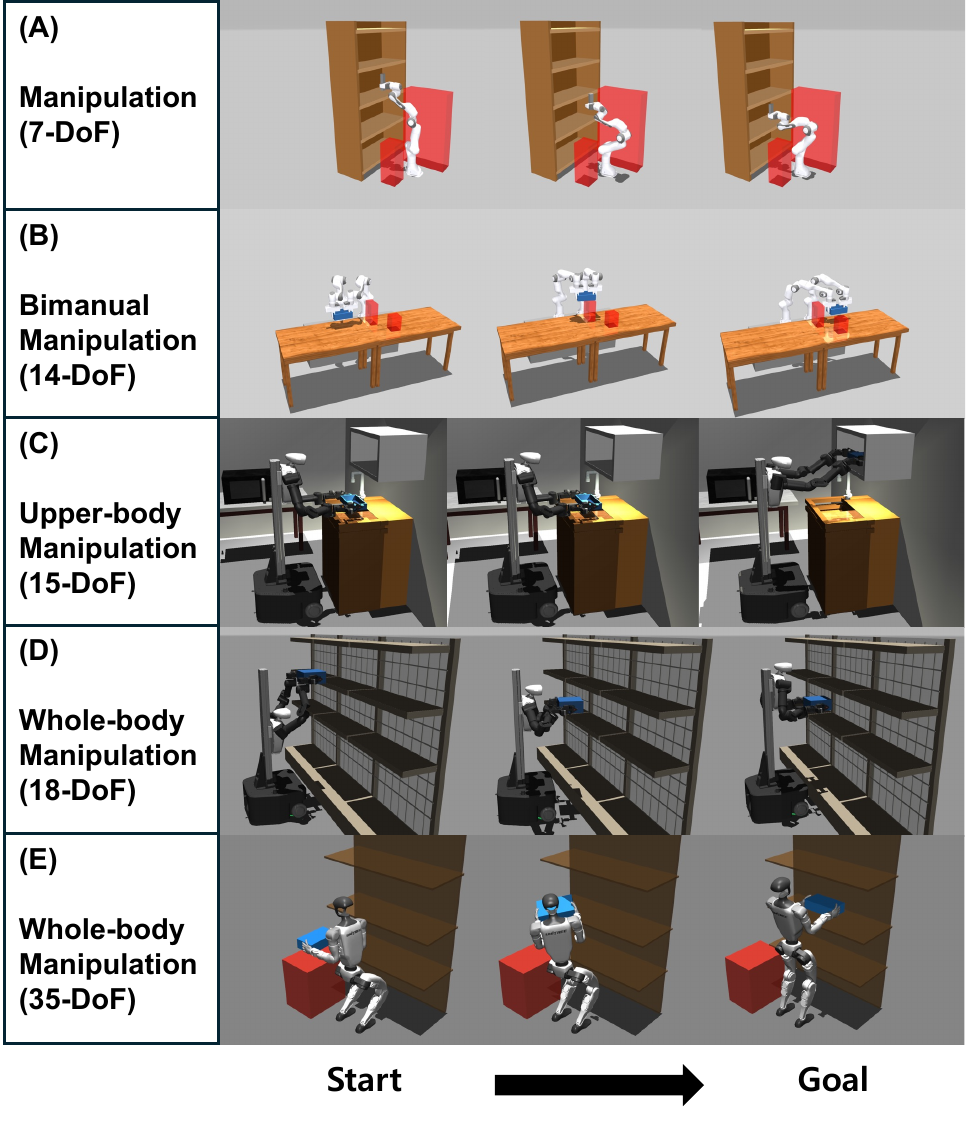}
    \vspace{-4mm}
    \captionsetup{font=small}
    \caption{
    Experimental setups used for constrained motion-planning benchmarks.
    Start and goal configurations are shown for the Franka Panda, dual Franka,
    fixed- and mobile-base FFW-SG2, and Unitree G1.
    }
    \label{fig:experiment_setup}
    \vspace{0mm}
\end{figure}

\begin{table}[t]
\centering

\captionsetup{font=footnotesize}

\caption{
Planning performance over 100 trials.\\
Time: mean $\pm$ std. over successful trials.
}
\label{tab:overall_performance}

\scriptsize
\renewcommand{\arraystretch}{1.0}
\setlength{\tabcolsep}{2.2pt}

\begin{tabular*}{\columnwidth}{
@{}
>{\raggedright\arraybackslash}p{0.1\columnwidth}
@{\hspace{2.2pt}\vrule width 0.2pt\hspace{2.2pt}}
>{\raggedright\arraybackslash}p{0.27\columnwidth}
@{\extracolsep{\fill}}
l
@{\hspace{2\tabcolsep}\extracolsep{0pt}}
c
r
@{}
}
\hline
\textbf{Robot}
&
\textbf{Constraint}
&
\textbf{Planner}
&
\textbf{Time (ms)}
&
\textbf{Succ.}
\\
\hline

\multirow{4}{*}{\textbf{Franka}}
&
\multirow{4}{*}{OR(2D)}
& TB-RRT
& 570.58 $\pm$ 446.23
& 98 \\

&
& McVAMP
& 5.20 $\pm$ 3.37
& 100 \\

&
& \cellcolor{gray!30}\textbf{cpRRTC}
& \cellcolor{gray!30}\textbf{0.52 $\pm$ 0.06}
& \cellcolor{gray!30}\textbf{100} \\

&
& PATACON
& 1.65 $\pm$ 0.10
& 100 \\
\hline

\multirow{6}{*}{\textbf{Dual}}
&
\multirow{3}{*}{BK(6D)}
& McVAMP
& 44.60 $\pm$ 23.69
& 100 \\

&
& \cellcolor{gray!30}\textbf{cpRRTC}
& \cellcolor{gray!30}\textbf{3.94 $\pm$ 0.39}
& \cellcolor{gray!30}\textbf{100} \\

&
& PATACON
& 4.77 $\pm$ 0.78
& 100 \\
\cline{2-5}

&
\multirow{3}{*}{BK + OR (8D)}
& McVAMP
& 270.49 $\pm$ 312.54
& 100 \\

&
& \cellcolor{gray!30}\textbf{cpRRTC}
& \cellcolor{gray!30}\textbf{5.55 $\pm$ 0.66}
& \cellcolor{gray!30}\textbf{100} \\

&
& PATACON
& 5.82 $\pm$ 0.76
& 100 \\
\hline

\multirow{4}{*}{\textbf{FFW-F}}
&
\multirow{2}{*}{BK (6D)}
& cpRRTC
& 29.49 $\pm$ 3.52
& 100 \\

&
& \cellcolor{gray!30}\textbf{PATACON}
& \cellcolor{gray!30}\textbf{20.95 $\pm$ 3.20}
& \cellcolor{gray!30}\textbf{100} \\
\cline{2-5}

&
\multirow{2}{*}{BK + OR (8D)}
& cpRRTC
& 291.96 $\pm$ 81.03
& 100 \\

&
& \cellcolor{gray!30}\textbf{PATACON}
& \cellcolor{gray!30}\textbf{25.94 $\pm$ 4.06}
& \cellcolor{gray!30}\textbf{100} \\
\hline

\multirow{2}{*}{\textbf{FFW-M}}
&
\multirow{2}{*}{BK + OR~/~CoM (8D/2D)}
& cpRRTC
& 117.79 $\pm$ 16.84
& 100 \\

&
& \cellcolor{gray!30}\textbf{PATACON}
& \cellcolor{gray!30}\textbf{60.14 $\pm$ 4.79}
& \cellcolor{gray!30}\textbf{100} \\
\hline

\multirow{4}{*}{\textbf{G1}}
&
\multirow{2}{*}{BK + FT~/~CoM (18D/2D)}
& cpRRTC
& 72.41 $\pm$ 19.25
& 100 \\

&
& \cellcolor{gray!30}\textbf{PATACON}
& \cellcolor{gray!30}\textbf{57.44 $\pm$ 9.00}
& \cellcolor{gray!30}\textbf{100} \\
\cline{2-5}

&
\multirow{2}{*}{BK + OR + FT~/~CoM (20D/2D)}
& cpRRTC
& 242.48 $\pm$ 19.47
& 100 \\

&
& \cellcolor{gray!30}\textbf{PATACON}
& \cellcolor{gray!30}\textbf{152.95 $\pm$ 44.29}
& \cellcolor{gray!30}\textbf{100} \\
\hline

\end{tabular*}

\vspace{1mm}

\begin{flushleft}
\tiny
FFW-F: FFW-SG2 with fixed base;
FFW-M: FFW-SG2 with mobile base.\\
BK: 6D, OR: 2D, FT: 12D, CoM (2D Inequality Constraints)\\
See Sec.~\ref{sec:experiments} for constraint definitions.
\end{flushleft}

\vspace{-5mm}
\end{table}

\begin{figure}[t]
\vspace{1mm}
    \centering
    \includegraphics[
    width=\columnwidth,
    trim=110 0 100 0,
    clip
    ]{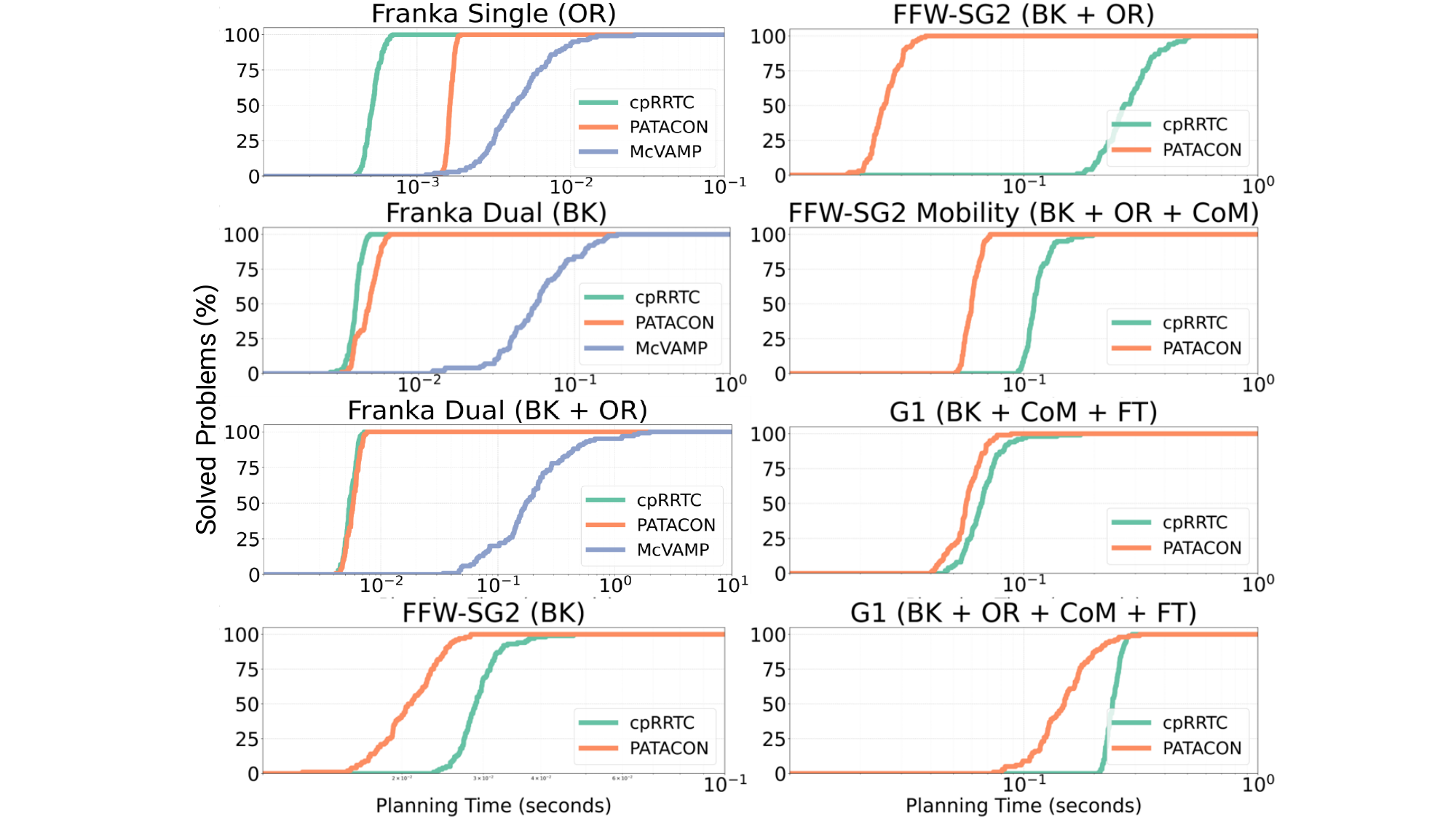}
    \caption{
    Cumulative planning success across constrained planning benchmarks.
    Each curve shows the percentage of 100 randomly sampled start--goal problems solved within the corresponding planning time.
    Planning time is shown on a logarithmic scale.
    }
    \label{fig:compare algorithm}
    \vspace{-3mm}
\end{figure}

On the Franka benchmarks, cpRRTC and PATACON show comparable planning performance, with their relative advantage varying across the constraint settings. In contrast, PATACON consistently achieves lower planning latency across the higher-dimensional FFW-SG2 and Unitree G1 benchmarks. Its relative advantage becomes more pronounced in the more restrictive constraint settings, particularly when the axis constraint is introduced. This trend is most evident on the fixed-base FFW-SG2 benchmark with bimanual-pose and axis constraints, where the mean planning latency corresponds to an equivalent planning throughput of approximately 38.5 solutions per second for PATACON, compared with 3.4 for cpRRTC. A similar tendency is observed on the Unitree G1 benchmarks, where PATACON is faster both with and without the axis constraint, while the performance gap increases after the axis constraint is added. These results suggest that the benefit of tangent-space-guided parallel exploration becomes more pronounced in higher-dimensional and more restrictive constrained planning regimes.

\subsection{Ablation Study}
\label{sec:ablation_study}

Starting from cpRRTC, +TGS introduces the tangent-guided exploration described in Sec.~\ref{sec:concurrent_exploration}. +TGMP further generates multiple consecutive candidate edges within a single EXTEND call, but processes them edge by edge: each candidate edge is projected and collision-validated before the next edge is processed. PATACON retains this tangent-guided multi-edge expansion, but projects and validates multiple candidate edges concurrently on the GPU while preventing backtracking toward previously explored tangent-space regions.

TGS alone increases planning time because constructing and maintaining tangent-space information introduces additional overhead despite providing geometry-aware exploration and a restricted nearest-neighbor search. TGMP reduces this overhead by reusing the same tangent-space context across multiple consecutive expansion steps, thereby amortizing the cost of tangent-space construction.

PATACON further reduces planning time by projecting and validating multiple candidate edges in parallel on the GPU. In addition, the backtracking-prevention strategy biases expansion away from the parent tangent-space region, reducing redundant exploration of already visited regions and promoting progress toward unexplored portions of the constraint manifold. Overall, the results show that tangent-guided exploration becomes most effective when combined with multi-edge expansion, GPU-parallel edge processing, and directed exploration that avoids backtracking.

\begin{table}[t]
\centering
\caption{
Cumulative ablation on the G1 whole-body benchmark.
}
\label{tab:ablation_results}
\captionsetup{font=small}
\renewcommand{\arraystretch}{1.05}
\setlength{\tabcolsep}{4pt}

\begin{tabular}{l c c}
\hline
\textbf{Variant}
&
\textbf{Time (ms) $\downarrow$}
&
\textbf{Success (\%) $\uparrow$}
\\
\hline

cpRRTC
& 242.48 $\pm$ 19.47
& 100 \\

+TGS
& 339.78 $\pm$ 117.54
& 100 \\

+TGMP
& 288.43 $\pm$ 123.17 
& 100 \\

\cellcolor{gray!30}\textbf{PATACON}
& \cellcolor{gray!30}\textbf{152.95 $\pm$ 44.29}
& \cellcolor{gray!30}\textbf{100} \\

\hline
\end{tabular}
\vspace{-5mm}
\end{table}

\subsection{Anytime Planning Performance}
\label{sec:anytime_performance}

The previous experiments demonstrate that PATACON can rapidly obtain a feasible solution. This low first-solution latency leaves additional computation time that can be exploited for further solution refinement. We therefore embed PATACON in an AORRTC-based anytime optimization framework~\cite{AORRTC} and evaluate whether the remaining computation budget can be used to progressively improve solution quality on the G1 whole-body benchmark.

The resulting anytime planner is denoted AO-PATACON. AO-PATACON does not continue expanding the existing trees after finding the first feasible solution. Instead, it uses the cost of that solution as an upper bound, reinitializes the trees, and restarts the search. Whenever a lower-cost solution is found, the cost bound is tightened and the search is restarted again. This process is repeated throughout the 10-s optimization budget.

The best-so-far solution cost at optimization time $t$ is defined as
\begin{equation}
C^{*}(t)
=
\min_{\tau \leq t}
C(\sigma_{\tau}),
\label{eq:anytime_cost}
\end{equation}
where $\sigma_{\tau}$ denotes a feasible solution obtained at time $\tau$, and the configuration-space path length is used as the solution cost.

To quantify the improvement obtained after the first feasible solution, we define the relative cost reduction as
\begin{equation}
R_{\mathrm{cost}}
=
\frac{
C_{\mathrm{init}}
-
C^{*}(T_{\mathrm{AO}})
}{
C_{\mathrm{init}}
}
\times 100,
\label{eq:anytime_reduction}
\end{equation}
where $C_{\mathrm{init}}$ denotes the cost of the first feasible solution.

\begin{figure}[t]
    \vspace{3mm}
    \centering
    \includegraphics[width=\columnwidth]{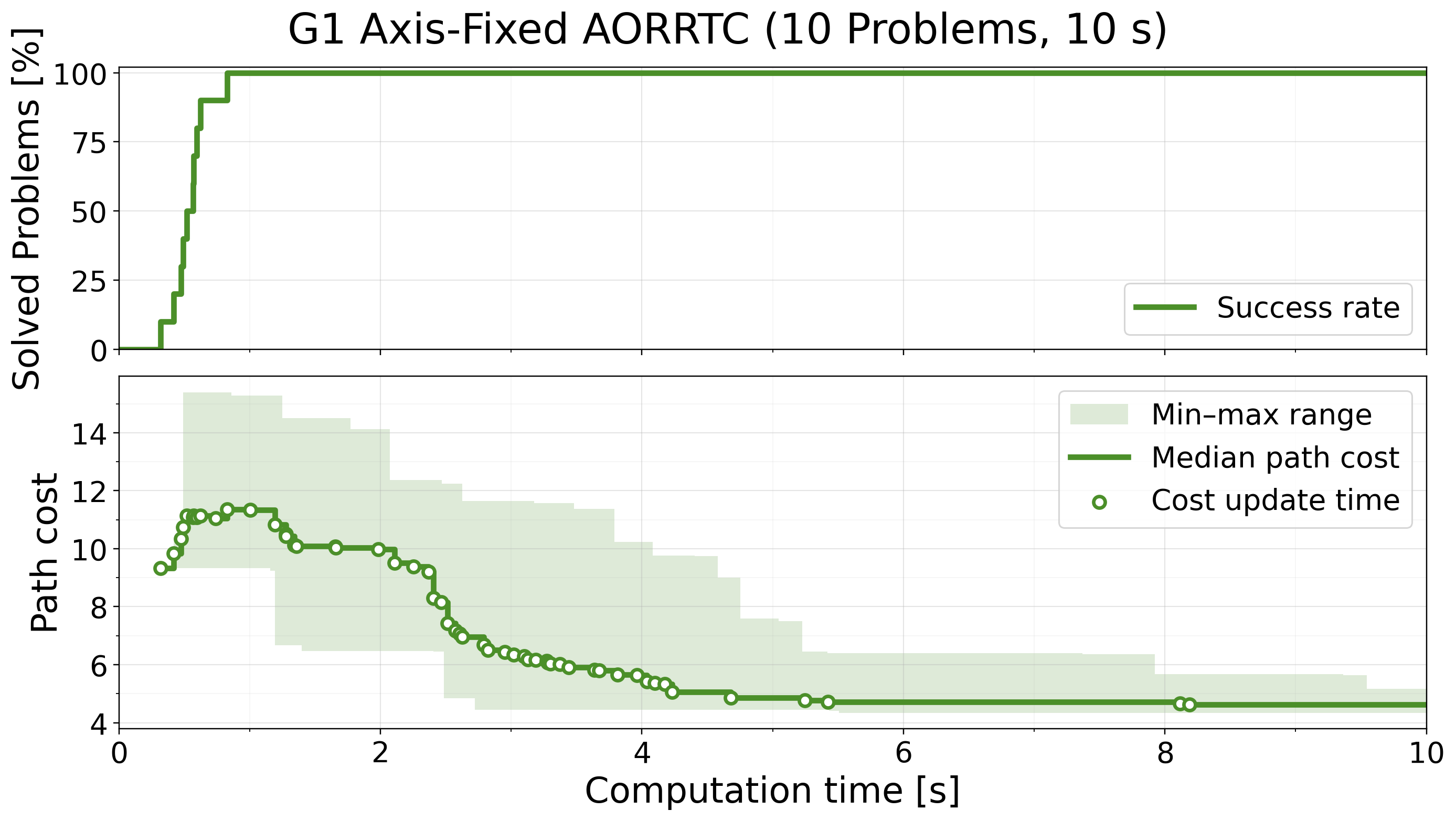}
    \caption{Anytime optimization behavior of AO-PATACON on the G1 whole-body benchmark over 10 independent runs. The upper plot shows the success rate over computation time, and the lower plot shows the median best-so-far path cost during continued optimization.}
    \label{fig:anytime_cost}
\end{figure}

As shown in Fig.~\ref{fig:anytime_cost}, AO-PATACON reaches a 100\% success rate during the early stage of planning and subsequently continues to improve the quality of the incumbent solution within the remaining optimization budget. The best-so-far path cost decreases in a stepwise manner as lower-cost feasible paths are discovered, with the largest reductions occurring shortly after the first feasible solutions are obtained. The magnitude of subsequent improvements gradually decreases, indicating convergence toward lower-cost solutions as additional computation time is provided.

These results show that, because PATACON finds the first feasible solution quickly, a larger portion of the fixed planning budget can be devoted to subsequent solution refinement. AO-PATACON can then use the remaining computation time to progressively improve path quality in an anytime manner.

\section{Conclusion}
\label{sec:conclusion}

In this work, we presented PATACON, a GPU-parallel tangent-bundle RRT-Connect framework for constrained motion planning. PATACON integrates tangent-space-guided exploration with GPU-parallel tree expansion and validation, enabling efficient exploration of constrained configuration spaces. Experimental results across dual-arm and whole-body planning problems show that PATACON becomes increasingly advantageous as the constraints become more restrictive and the planning problem becomes more challenging. These results demonstrate the effectiveness of combining local tangent-space guidance with massively parallel GPU execution for high-dimensional constrained motion planning.

\section*{Acknowledgments}
We used ChatGPT (OpenAI) to refine the language of
the manuscript and Codex (OpenAI) to assist in writing
portions of the code used in the experiments reported
in Section~V.

\appendix[Correction of Benchmark Results]

After the initial submission, we found that
a computational task had not terminated and
was still running on the same GPU during
benchmarking. This caused resource contention
and led to overestimated cpRRTC planning times
in the Franka OR (2D) and Dual Franka BK+OR (8D)
benchmarks.
The originally reported planning times of
$2.11 \pm 2.18$ ms and $8.94 \pm 4.90$ ms
were corrected to $0.52 \pm 0.06$ ms and
$5.55 \pm 0.66$ ms, respectively, after
re-evaluation without concurrent GPU workloads.
The corrected results, reflected in Table~I
and Fig.~5, show that cpRRTC achieves lower
mean planning times than PATACON in both
benchmarks.

\bibliographystyle{IEEEtran}
\bibliography{planning_papers}

@inproceedings{RRTConnect,
author    = {Kuffner, James J. and LaValle, Steven M.},
title     = {{RRT}-Connect: An Efficient Approach to Single-Query Path Planning},
booktitle = {Proceedings of the IEEE International Conference on Robotics and Automation (ICRA)},
volume    = {2},
pages     = {995--1001},
year      = {2000},
doi       = {10.1109/ROBOT.2000.844730}
}

@article{KingstonConstraintSurvey,
author    = {Kingston, Zachary and Moll, Mark and Kavraki, Lydia E.},
title     = {Sampling-Based Methods for Motion Planning with Constraints},
journal   = {Annual Review of Control, Robotics, and Autonomous Systems},
volume    = {1},
pages     = {159--185},
year      = {2018},
doi       = {10.1146/annurev-control-060117-105226}
}

@article{StilmanTaskConstraint,
author    = {Stilman, Mike},
title     = {Global Manipulation Planning in Robot Joint Space with Task Constraints},
journal   = {IEEE Transactions on Robotics},
volume    = {26},
number    = {3},
pages     = {576--584},
year      = {2010},
doi       = {10.1109/TRO.2010.2044949}
}

@inproceedings{CBiRRT,
author    = {Berenson, Dmitry and Srinivasa, Siddhartha S. and Ferguson, Dave and Kuffner, James J.},
title     = {Manipulation Planning on Constraint Manifolds},
booktitle = {Proceedings of the IEEE International Conference on Robotics and Automation (ICRA)},
pages     = {625--632},
year      = {2009},
doi       = {10.1109/ROBOT.2009.5152399}
}

@article{TaskSpaceRegions,
author    = {Berenson, Dmitry and Srinivasa, Siddhartha S. and Kuffner, James J.},
title     = {Task Space Regions: A Framework for Pose-Constrained Manipulation Planning},
journal   = {The International Journal of Robotics Research},
volume    = {30},
number    = {12},
pages     = {1435--1460},
year      = {2011},
doi       = {10.1177/0278364910396389}
}

@article{AtlasRRT,
author    = {Jaillet, L{\'e}onard and Porta, Josep M.},
title     = {Path Planning under Kinematic Constraints by Rapidly Exploring Manifolds},
journal   = {IEEE Transactions on Robotics},
volume    = {29},
number    = {1},
pages     = {105--117},
year      = {2013},
doi       = {10.1109/TRO.2012.2222272}
}

@article{TBRRT,
author    = {Kim, Beobkyoon and Um, Terry Taewoong and Suh, Chansu and Park, Frank C.},
title     = {Tangent Bundle {RRT}: A Randomized Algorithm for Constrained Motion Planning},
journal   = {Robotica},
volume    = {34},
number    = {1},
pages     = {202--225},
year      = {2016},
doi       = {10.1017/S0263574714001234}
}

@article{KingstonIMACS,
author    = {Kingston, Zachary and Moll, Mark and Kavraki, Lydia E.},
title     = {Exploring Implicit Spaces for Constrained Sampling-Based Planning},
journal   = {The International Journal of Robotics Research},
volume    = {38},
number    = {10--11},
pages     = {1151--1178},
year      = {2019},
doi       = {10.1177/0278364919868530}
}

@article{Ichter2019Latent,
author    = {Ichter, Brian and Pavone, Marco},
title     = {Robot Motion Planning in Learned Latent Spaces},
journal   = {IEEE Robotics and Automation Letters},
volume    = {4},
number    = {3},
pages     = {2407--2414},
year      = {2019},
doi       = {10.1109/LRA.2019.2901898}
}

@article{ParkLatentMotion,
author    = {Park, Suhan and Jeon, Suhyun and Park, Jaeheung},
title     = {A Constrained Motion Planning Method Exploiting Learned Latent Space for High-Dimensional State and Constraint Spaces},
journal   = {IEEE/ASME Transactions on Mechatronics},
volume    = {29},
number    = {4},
pages     = {3001--3009},
year      = {2024},
doi       = {10.1109/TMECH.2024.3399594}
}

@article{CoMPNetX,
author    = {Qureshi, Ahmed Hussain and Dong, Jiangeng and Baig, Asfiya and Yip, Michael C.},
title     = {Constrained Motion Planning Networks {X}},
journal   = {IEEE Transactions on Robotics},
volume    = {38},
number    = {2},
pages     = {868--886},
year      = {2022},
doi       = {10.1109/TRO.2021.3096070}
}

@inproceedings{BialkowskiParallelRRT,
author    = {Bialkowski, Joshua and Karaman, Sertac and Frazzoli, Emilio},
title     = {Massively Parallelizing the {RRT} and the {RRT*}},
booktitle = {Proceedings of the IEEE/RSJ International Conference on Intelligent Robots and Systems (IROS)},
pages     = {3513--3518},
year      = {2011},
doi       = {10.1109/IROS.2011.6095053}
}

@article{CForest,
author    = {Otte, Michael and Correll, Nikolaus},
title     = {{C-FOREST}: Parallel Shortest-Path Planning with Superlinear Speedup},
journal   = {IEEE Transactions on Robotics},
volume    = {29},
number    = {3},
pages     = {798--806},
year      = {2013},
doi       = {10.1109/TRO.2013.2240176}
}

@inproceedings{cuRobo,
author    = {Sundaralingam, Balakumar and Hari, Siva Kumar Sastry and Fishman, Adam and Garrett, Caelan and Van Wyk, Karl and Blukis, Valts and Millane, Alexander and Oleynikova, Helen and Handa, Ankur and Ramos, Fabio and Ratliff, Nathan and Fox, Dieter},
title     = {{CuRobo}: Parallelized Collision-Free Robot Motion Generation},
booktitle = {Proceedings of the IEEE International Conference on Robotics and Automation (ICRA)},
pages     = {8112--8119},
year      = {2023},
doi       = {10.1109/ICRA48891.2023.10160765}
}

@inproceedings{VAMP,
author    = {Thomason, Wil and Kingston, Zachary and Kavraki, Lydia E.},
title     = {Motions in Microseconds via Vectorized Sampling-Based Planning},
booktitle = {Proceedings of the IEEE International Conference on Robotics and Automation (ICRA)},
pages     = {8749--8756},
year      = {2024},
doi       = {10.1109/ICRA57147.2024.10611190}
}

@inproceedings{McVAMP,
author    = {Iyer, Shrutheesh R. and Chang, I-Chia and Liu, Andrew Z. and Gu, Yan and Kingston, Zachary},
title     = {Vectorizing Projection in Manifold-Constrained Motion Planning for Real-Time Whole-Body Control},
booktitle = {Proceedings of the IEEE/RSJ International Conference on Intelligent Robots and Systems (IROS)},
year      = {2026},
note      = {To Appear}
}

@inproceedings{pRRTC,
author    = {Huang, Chih H. and Jadhav, Pranav and Plancher, Brian and Kingston, Zachary},
title     = {{pRRTC}: {GPU}-Parallel {RRT}-Connect for Fast, Consistent, and Low-Cost Motion Planning},
booktitle = {Proceedings of the IEEE International Conference on Robotics and Automation (ICRA)},
year      = {2026}
}

@misc{cpRRTC,
author        = {Hu, Jiaming and Wang, Jiawei and Christensen, Henrik I.},
title         = {{cpRRTC}: {GPU}-Parallel {RRT}-Connect for Constrained Motion Planning},
year          = {2025},
eprint        = {2505.06791},
archivePrefix = {arXiv},
primaryClass  = {cs.RO}
}

@article{BatchRRT,
author    = {Liu, Jie and Li, Meng and Liu, Zian and Sun, Tianmei},
title     = {{BatchRRT}: A Fully-Batched Reformulation of {RRT} with {GPU}-Accelerated {SDF} Collision Checking for Real-Time Human-Robot Collision Avoidance},
journal   = {IEEE Robotics and Automation Letters},
volume    = {11},
number    = {5},
pages     = {6384--6391},
year      = {2026},
doi       = {10.1109/LRA.2026.3681160}
}

@article{Halton1960,
author    = {Halton, J. H.},
title     = {On the Efficiency of Certain Quasi-Random Sequences of Points in Evaluating Multi-Dimensional Integrals},
journal   = {Numerische Mathematik},
volume    = {2},
number    = {1},
pages     = {84--90},
year      = {1960},
doi       = {10.1007/BF01386213}
}

@inproceedings{BranickyQuasiRandom,
author    = {Branicky, Michael S. and LaValle, Steven M. and Olson, Kari and Yang, Libo},
title     = {Quasi-Randomized Path Planning},
booktitle = {Proceedings of the IEEE International Conference on Robotics and Automation (ICRA)},
volume    = {2},
pages     = {1481--1487},
year      = {2001},
doi       = {10.1109/ROBOT.2001.932820}
}

@article{YershovaNN,
author    = {Yershova, Anna and LaValle, Steven M.},
title     = {Improving Motion-Planning Algorithms by Efficient Nearest-Neighbor Searching},
journal   = {IEEE Transactions on Robotics},
volume    = {23},
number    = {1},
pages     = {151--157},
year      = {2007},
doi       = {10.1109/TRO.2006.886840}
}

@article{YakeyClosedChain,
author    = {Yakey, John H. and LaValle, Steven M. and Kavraki, Lydia E.},
title     = {Randomized Path Planning for Linkages with Closed Kinematic Chains},
journal   = {IEEE Transactions on Robotics and Automation},
volume    = {17},
number    = {6},
pages     = {951--958},
year      = {2001}
}

@inproceedings{IchnowskiParallel,
author    = {Ichnowski, Jeffrey and Alterovitz, Ron},
title     = {Parallel Sampling-Based Motion Planning with Superlinear Speedup},
booktitle = {Proceedings of the IEEE/RSJ International Conference on Intelligent Robots and Systems (IROS)},
pages     = {1206--1212},
year      = {2012}
}

@inproceedings{JacobsParallelSBMP,
author    = {Jacobs, Steven A. and Manavi, Kaveh and Burgos, Javier and Denny, Jory and Thomas, Shawna and Amato, Nancy M.},
title     = {A Scalable Method for Parallelizing Sampling-Based Motion Planning Algorithms},
booktitle = {Proceedings of the IEEE International Conference on Robotics and Automation (ICRA)},
pages     = {2529--2536},
year      = {2012}
}

@article{PanGPUCollision,
author    = {Pan, Jia and Manocha, Dinesh},
title     = {{GPU}-Based Parallel Collision Detection for Fast Motion Planning},
journal   = {The International Journal of Robotics Research},
volume    = {31},
number    = {2},
pages     = {187--200},
year      = {2012}
}

@incollection{KleinbortBottleneck,
author    = {Kleinbort, Michal and Salzman, Oren and Halperin, Dan},
title     = {Collision Detection or Nearest-Neighbor Search? On the Computational Bottleneck in Sampling-Based Motion Planning},
booktitle = {Algorithmic Foundations of Robotics XII},
publisher = {Springer},
pages     = {624--639},
year      = {2020}
}

@techreport{BalancedRRT,
author      = {Kuffner, James J. and LaValle, Steven M.},
title       = {An Efficient Approach to Path Planning Using Balanced Bidirectional {RRT} Search},
institution = {Robotics Institute, Carnegie Mellon University},
address     = {Pittsburgh, PA},
year        = {2005}
}

@article{AORRTC,
author    = {Wilson, Tyler S. and Thomason, Wil and Kingston, Zachary and Gammell, Jonathan D.},
title     = {{AORRTC}: Almost-Surely Asymptotically Optimal Planning with {RRT}-Connect},
journal   = {IEEE Robotics and Automation Letters},
year      = {2025},
note      = {Accepted September 2025}
}

@inproceedings{Eqnet2021,
  title={Learning equality constraints for motion planning on manifolds},
  author={Sutanto, Giovanni and Fern{\'a}ndez, Isabel Rayas and Englert, Peter and Ramachandran, Ragesh Kumar and Sukhatme, Gaurav},
  booktitle={Conference on Robot Learning},
  pages={2292--2305},
  year={2021},
  organization={PMLR}
}

\end{document}